\documentclass[10pt,twocolumn,letterpaper]{article}

\usepackage{wacv}
\usepackage{times}
\usepackage{epsfig}
\usepackage{graphicx}
\usepackage[cmex10]{amsmath}
\usepackage{amssymb}
\usepackage{algorithmic}
\usepackage{subfigure}
\usepackage{algorithm}
\usepackage{cite}
\usepackage{graphicx}
\usepackage{url}

\usepackage[font=it]{caption}
\usepackage{tikz}

\wacvfinalcopy 

\def\wacvPaperID{376} 

\ifwacvfinal\fi
\begin{document}

\title{Can we still avoid automatic face detection?}

\author{
Michael J. Wilber$^{1,2}$
$\qquad$
Vitaly Shmatikov$^{1,2}$
$\qquad$
Serge Belongie$^{1,2}$\\
\\
$^1$ Department of Computer Science, Cornell University\quad$^2$ 	Cornell Tech
}

\maketitle
\ifwacvfinal\thispagestyle{empty}\fi
\begin{abstract}
After decades of study, automatic face detection and recognition systems are now accurate and widespread. Naturally, this means users who wish to avoid automatic recognition are becoming less able to do so. Where do we stand in this cat-and-mouse race? We currently live in a society where everyone carries a camera in their pocket. Many people willfully upload most or all of the pictures they take to social networks which invest heavily in automatic face recognition systems. In this setting, is it still possible for privacy-conscientious users to avoid automatic face detection and recognition? If so, how? Must evasion techniques be obvious to be effective, or are there still simple measures that users can use to protect themselves?

In this work, we find ways to evade face detection on Facebook, a representative example of a popular social network that uses automatic face detection to enhance their service. We challenge widely-held beliefs about evading face detection: do our old techniques such as blurring the face region or wearing ``privacy glasses'' still work? We show that in general, state-of-the-art detectors can often find faces even if the subject wears occluding clothing or even if the uploader damages the photo to prevent faces from being detected.
\end{abstract}
\begin{figure}[tp]
\centering
\includegraphics[width=\linewidth]{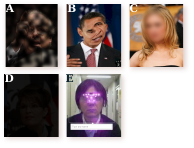}
\caption{\label{fig:hardexamples}Can you see the face in each of these images? \textbf{Facebook's automatic face detector can detect and localize all six faces shown above}, even when we try to hide the face by (A)~adding occluding noise, (B)~adding distortion, (C)~blurring the face region, or (D)~altering image lighting. Deliberate countermeasures such as (E)~wearing a ``privacy visor'' with infrared LEDs~\cite{yamada_privacy_2013} are not always effective---sometimes Facebook can see through these disguises too. Facebook may not be able to \emph{recognize} these faces, but if Facebook can \emph{detect} them, it may prompt friends to tag these hard faces and reveal their identity. }
\end{figure}

\section{Introduction and background}
``Please find all the pictures that my friends took of me.'' Thanks to clever tools from large social networks like Facebook and Google Photos, it is now possible to answer this question with a high degree of accuracy.
%
Face recognition is a \emph{de facto} flagship example of applied computer vision. 
In fact, Facebook's recognition system is said to approach human-level accuracy under controlled conditions~\cite{taigman_deepface:_2013}. Until very recently, it was considered state-of-the-art on the well-accepted ``Labeled Faces in the Wild (LFW)'' face verification benchmark, beating dozens of other academic and commercial systems in the unrestricted setting~\cite{lfw-results}. Automatically annotating a photo album with this rich structure ``for free'' is a big help when compiling scrapbooks, searching for vacation photos, or catching up on a friend's whereabouts. In particular, Facebook's recently-announced ``Photo Magic'' app will use face recognition to help users quickly share photos among tagged friends~\cite{goldman_facebook_2015}.

However, this accuracy creates new and interesting power dynamics that remain largely unexplored because automatic face recognition technology is actively harmful for many users. Consider a middle school student fighting Internet bullying, a member of a witness protection program, a whistleblower hiding from law enforcement, or an activist protesting their local government.
As face recognition systems grow more accurate, these people lose the ability to hide their identity, for better or worse. As we will see---and as hinted in Fig.~\ref{fig:hardexamples}---Facebook's face detector can detect (but not necessarily recognize) faces that are difficult for humans to find, even when steps are taken to prevent detection. These issues stretch beyond Facebook: any accurate biometric system shares similar trade-offs~\cite{tboult}. Many privacy advocates argue that legitimate uses of automatic face recognition and similar biometrics should be carefully considered and closely regulated.


The options are limited for those seeking refuge from automatic recognition systems. Clearly, the best way to ``opt out'' is to never use online social networking tools again, but it can be difficult and lonesome to stop using Facebook completely. Further, even if a person does not have an account on Facebook, their Facebook-using friends can still ``tag'' their name in pictures. In these cases, Facebook could potentially build a recognition model anyway without exposing it to other users. Further, while Facebook provides tools for users to ``opt-out'' of automatic tagging, this action does not stop Facebook from merely hiding recognition results until law enforcement or a malicious user discovers how to uncover them. It is always up to the user to take steps to protect their own privacy.

Another way to evade automatic face recognition is to damage the photograph so badly that the face can no longer be automatically recognized or detected. Needless to say, this is not very practical. If a user wishes to avoid recognition this way, \emph{any friends who might take a picture of her} must promise never to tag her and must always damage all photos they ever take of her. The incentives are misaligned because the photographer certainly does not want to damage the photograph and may not understand the subject's privacy situation.

These two perspectives--uploader and subject--suggest complimentary strategies to evade face detection. In this work, we investigate both angles. Acting as the photographer, we explore the space of possible synthetic image transformations using various image filters to foil face detection without degrading image quality. Second, as the subject, we investigate whether wearing sunglasses, scarves, hats, or other occluding clothing is an effective way to avoid face detection.
These approaches are interesting to us for two reasons: first, this work illuminates the accuracy of Facebook's face detector \emph{today} and may uncover simple techniques that journalists, whistleblowers, and other privacy-conscious users can use to perturb photos of themselves. Second, this study illustrates the gap between automatic and human face detection performance.

Our main contribution is a detailed case study showing the accuracy of Facebook's 
face detector after applying several image transformations. What techniques, if any, might allow users to foil face detection and recognition in photos they upload? Do existing countermeasures such as censor bars, ``privacy glasses,'' and image blurring withstand the test of time?

Though our results are specific to Facebook's face detector, we emphasize that they are not the main focus of our study. Our results could generalize to other commonly-used face detectors such as the one powering face search in \emph{Google Photos} or Snapchat's \emph{Lenses} feature.

\section{Background}
\subsection{Face recognition in general}
In order to foil face recognition, we must first consider the various parts of a typical face recognition pipeline. Consider the ``Social Network'' setting in which a user takes a picture of their family having lunch in a park. After uploading this picture, the social network automatically \emph{tags} the people that appear in the user's picture. This photo is then shared with their friends. Depending on the user's settings, it may automatically become searchable by date, time, place, or the name of the people who appear in the photo. How does this complicated process work? Traditional recognition systems typically use five steps:

\begin{itemize}
\item \textbf{Face detection}: Given a raw image, find bounding boxes for all of the candidate faces within that image. This is essentially an object detection or tracking problem~\cite{viola-facedetection,babenko-miltrack} and classic approaches use boosting in a multiple-instance learning framework to find potential faces. This step is our principal focus.
\item \textbf{Pose estimation}: For each candidate face region, estimate the location of several \emph{keypoints} such as the eyes, ears, nose, and mouth. This requires a system that is adept to changes in rotation, scale, translation, lighting, expression, and occlusion due to the inherent non-rigidity of the face and its environment~\cite{belhumeur_localization, burgos-artizzu_robust_2013}.
\item \textbf{Frontalization / normalization}: This step is designed to make the later stages of the pipeline invariant to pose/illumination/expression differences. To do this, a normalization step warps or maps the face image to some canonical reference image. This warp may be a 2D affine transform, a 2D piece-wise triangular warp~\cite{berg_tom-vs-pete_2012}, or a 3D camera re-projection~\cite{taigman_deepface:_2013}. Some systems additionally try to correct for lighting~\cite{wang_improving_2008, wang_face_2004} and expression~\cite{berg_tom-vs-pete_2012} differences.
\item \textbf{Feature extraction}: Once the face image is canonicalized, it is then represented in a well-defined vectorial representation. This may be as simple as taking pixel intensity values near the estimated keypoints~\cite{ren_face_2014}, or it may be based on convolution features acquired using deep learning~\cite{sun_deepid2,taigman_deepface:_2013} or semantically meaningful attributes~\cite{attribute_and_simile}.
\item \textbf{Recognition}: Once the face image is converted to some well-formed representation, many face tagging services compare the face to several stored models comprising a \emph{gallery} of the uploader's friends to find out who the face is. This step varies quite drastically between systems. Many \emph{face verification} systems simply compare the query face to each and every model in the gallery, asking ``Do these two faces belong to the same person?''~\cite{taigman_deepface:_2013}. Other \emph{face recognition} systems directly predict the face's identity using a multiclass classifier. It is unclear which approach Facebook uses in production, but they do evaluate one published system for comparison to the academic state-of-the-art in a verification setting~\cite{taigman_deepface:_2013}.
\end{itemize}
An additional implicit \textbf{enrollment} step is required, in which the user uploads images of themselves to train the model.

All of the steps of the pipeline are important for recognition success, and any of them may be subverted through various high-level actions. For example, occluding the face may cause detection to fail, warping the face in strange ways may cause pose estimation to fail, and introducing confusor identities or intentionally mis-tagging one's own images may cause the recognition step to incorrectly identify the face.

\begin{figure}[t]
\centering
\includegraphics[width=1\linewidth]{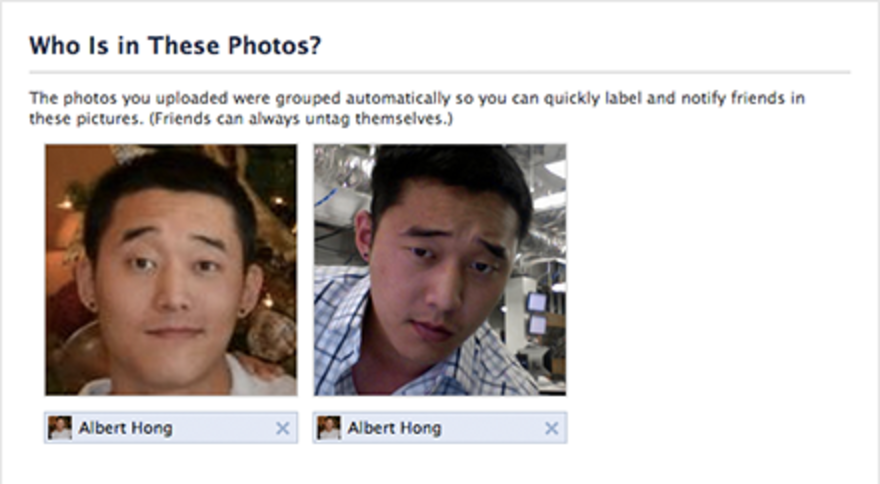}
\caption{\label{fig:tagface_prompt}Facebook can recover if part of the recognition pipeline fails. For instance, when the detection step finds a face but the recognition step fails, Facebook can prompt the uploader's friends to tag the people in the image, as shown in this screenshot. However, if the detection step does not find a face in the image, social networks will not generally prompt friends to tag the image. This is why we focus on the face detection step. This screenshot is from Facebook's ``Timeline and Tagging Settings'' page.}
\end{figure}

One practical way to break recognition is to subvert \emph{face detection}. We focus on this step for two reasons: first, automatic recognition systems cannot learn anything about faces that are impossible to detect. Since people take pictures without faces all the time, there is nothing strange about them. Second, if the detection step succeeds, some social networks may prompt the uploader's friends to tag the face even if the recognition fails, allowing one's friends to ``rat out'' the privacy-evading individual. A screenshot of this prompt taken from Facebook's \emph{Timeline and Tagging Settings} page is shown in \ref{fig:tagface_prompt}.

\subsection{Related work}
Many projects consider privacy enhancement tools that foil face detection by altering the actual face's appearance as seen by the imaging sensor. For example, consider the \emph{Privacy Visor} designed by Echizen~\etal \cite{yamada_privacy_2013}. This device consists of several high-powered infrared LEDs mounted on a pair of glasses. When activated, the infrared LEDs wash out the face when captured with conventional digital cameras, but are invisible to the human eye. The user's friends see a person wearing funny glasses, but their cameras only see specular highlights. 

Both of these approaches are practical and attractive because they allow the \emph{individual} to take steps to protect their \emph{own} privacy, without having to remind their friends to alter their photos. However, there are two principal drawbacks with these kinds of approaches: first, approaches based on pushing the user's appearance away from the average face only make the user \emph{more} iconic. No one wants to be ``that person with the funny glasses/makeup.'' Distinctive faces are easier for people to remember and recognize~\cite{sarno_attractiveness_1997,bruce_whats_1994}, which could work against the user's wishes to remain anonymous in the physical world. Second, neither countermeasure is always effective against Facebook's face detector. Fig.~\ref{fig:hardexamples} parts~(E) and~(F) shows screenshots of Facebook's image upload process. Though they may \emph{reduce} detection rates, some countermeasures such as Privacy Visor are ineffective for the example images we chose: Facebook detects the face and asks the user to tag the identity. Granted, Facebook could not \emph{always} find the face in either method, but only a few good detections are necessary to begin building a recognition model.

Many studies also investigate how image appearance affects face detection, albeit usually in a ``How can we make face detection better?'' sense. For example, Parris \etal~\cite{fedhd} organized a ``Face and Eye Detection on Hard Datasets'' challenge at IJCB~2011. Over a dozen commercial and academic contestants submitted face detection entries. The results reveal that state-of-the-art face detectors generally have trouble recognizing faces that are severely out-of-focus or small.

Other studies in this area include Scheirer \etal 's ``Face in the Branches'' detection task \cite{scheirer_perceptual_2014}. Scheirer varies the amount of occlusion of face images and compares detection accuracy between humans, Google Picasa, and Face.com. There is still a large gap in performance---human workers achieve 60\% face detection accuracy when as little as 25\% of the face is visible, but Picasa requires 60\% of the face to be visible for similar levels of accuracy. As we will show, we can exploit this difference to generate pictures that still look ``OK'' to humans but are unrecognizable by automatic systems. Finally,~\cite{juefei-xu_preliminary_2015} investigates one way of correcting occluded faces. If a forensic investigator needs a recognition result from an off-the-shelf recognition system, they can copy face parts from other faces over the occlusions, ``forcing'' the missing face to be detected. This manual correction step adversely affects recognition performance, but it demonstrates one potential way to counter the suggestions in our report.

\section{Synthetic experiment design}
\textbf{Facebook testing.} Facebook provides an extensive testing framework for app developers on the Facebook platform. For example, we can use Facebook's app development tools to create up to 2,000 \emph{sandbox test accounts}, which are exempt from Facebook's spam blocking and fake account detection policies. These test users cannot interact with anyone else on the main site, but other than this difference, test accounts are identical in form and functionality to real Facebook users: they can befriend other sandbox users, post messages, and upload photos to the site, either programmatically through the Open Graph API or from an authenticated browser session. At the time of writing, Facebook runs the face detection and recognition pipeline on photos uploaded by sandbox accounts, giving us a simple way to evaluate Facebook's recognition pipeline \emph{in vivo} without interfering with legitimate users.

After a sandbox user uploads an image, Facebook runs the face recognition pipeline. Detection results are exposed to the Javascript image viewer, so we used an authenticated headless browser session to extract these bounding boxes and potential recognition results of all detected faces. Though detection results are available instantly, it takes several days for Facebook to build a sandbox user's recognition model. No confidence score for either detection or recognition seems to be available. We only consider detection results in this report.

\textbf{Dataset.} We use the PubFig~\cite{attribute_and_simile} dataset, a very broad set containing images of 200 celebrities captured under real-world ``in-the-wild'' conditions. Each image is tagged with a single bounding box that describes the person's face region within the image. Unfortunately, to avoid copyright infringement, the authors of PubFig only distribute the URLs of each image and most of them are no longer accessible. Some URLs now point to incorrectly scaled versions of the original images or different images completely. All in all, we were only able to collect 9,195 of the 58,797 images. From this set, we uniformly sampled a single evaluation set of 100 images for use in all of our synthetic experiments.

Every image from Pubfig may contain multiple faces and Facebook may return multiple face bounding boxes for a given image. However, each image in Pubfig contains only one groundtruth bounding box. An image is considered a true accept if any of Facebook's boxes have an Intersection-over-Union (IoU) score greater than 10\% with the groundtruth.\footnote{The IoU score is defined as the area of the two boxes' intersection divided by the area of their union.} This conservative estimate is necessary because Pubfig's groundtruth bounding boxes are typically much tighter than those returned from Facebook's pipeline. Good facial keypoint extractors can often correct bad bounding boxes to a limited extent.

Though Pubfig was captured under unconstrained conditions, the faces are easy to detect. In fact, Facebook's face detector is generally more accurate than the groundtruth because of the non-static nature of Pubfig image URLs. Facebook outputs three ``false negatives'' in the unfiltered 100-image test set, but two of these three have incorrect groundtruth due to changes in the image URL since Pubfig was published. From manual inspection, Facebook only actually missed a single face in our evaluation set.

\textbf{Filters.} To study ways of evading Facebook's face detection, we designed several handcrafted image filters that cause Facebook's face detection to fail. We swept each filter through a set of appropriate parameters to measure performance with respect to the amount of distortion that each filter applies. In total, we uploaded more than 19,000 images to Facebook sandbox accounts.

\section{Selected synthetic results and discussion}
Unfortunately, most of our filters substantially degrade the picture's quality. As we increase the distortion, humans and machines alike can no longer see the faces. However, there are two categories of filters that have interesting performance characteristics: the \textbf{human-hurting} filters that measurably hurt human detection performance but do not affect Facebook, and the \textbf{machine-hurting} filters that hurt Facebook's detection ability even though the resulting image still looks reasonable. In this section, we describe some interesting filters from these two categories. The results are summarized in Tab.~\ref{tab:results}. All of the other filters are described in the supplementary material.
\begin{table}
\caption{\label{tab:results}Summary of our filters, approximate detection accuracies under the strongest settings, and subjective amount of image degradation.}
\begin{tabular}{lccl}
Filter & Accuracy & Degradation & \\
\hline
Noise & 0.22 & Low &
\parbox[c]{1cm}{\includegraphics[width=1cm]{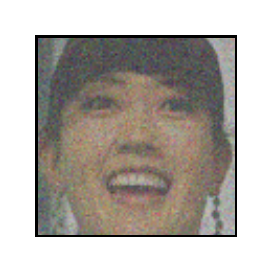}}\\
Blur & 0.37 & Med &
\parbox[c]{1cm}{\includegraphics[width=1cm]{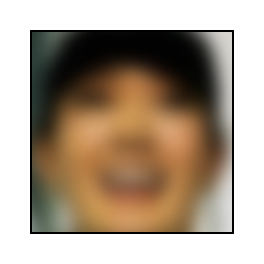}}\\
Blur Periocular & 0.30 & Low &
\parbox[c]{1cm}{\includegraphics[width=1cm]{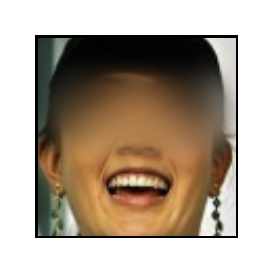}}\\
Darken & 0.94 & High &
\parbox[c]{1cm}{\includegraphics[width=1cm]{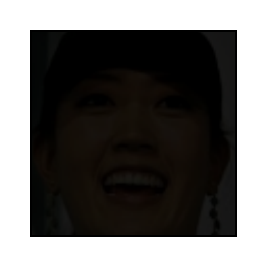}}\\
Censor (Black) & 0.21 & Med &
\parbox[c]{1cm}{\includegraphics[width=1cm]{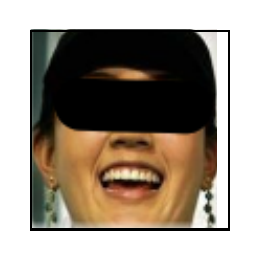}}\\
Censor (White) & 0.02 & Med &
\parbox[c]{1cm}{\includegraphics[width=1cm]{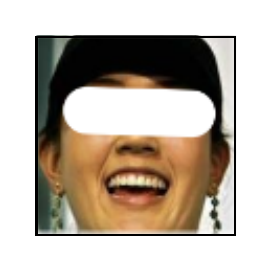}}\\
JPEG & 0.17 & Low &
\parbox[c]{1cm}{\includegraphics[width=1cm]{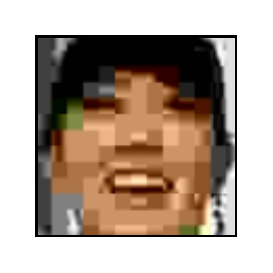}}\\
Leopard Spots & 0.01 & High &
\parbox[c]{1cm}{\includegraphics[width=1cm]{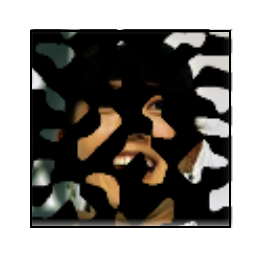}}\\
Rotation & 0.01 & High &
\parbox[c]{1cm}{\includegraphics[width=1cm]{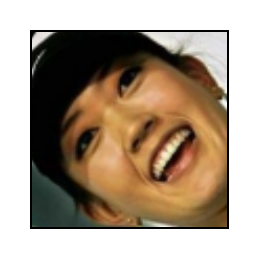}}\\
Swirl & 0.34 & High &
\parbox[c]{1cm}{\includegraphics[width=1cm]{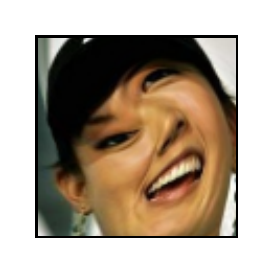}}\\
Warp & 0.01 & High &
\parbox[c]{1cm}{\includegraphics[width=1cm]{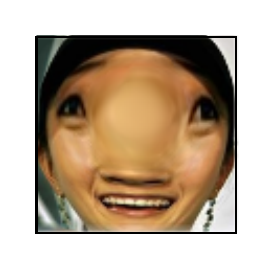}}
\end{tabular}
\end{table}

\subsection{Gaussian noise}
\begin{figure}[t]
\includegraphics[width=1\linewidth]{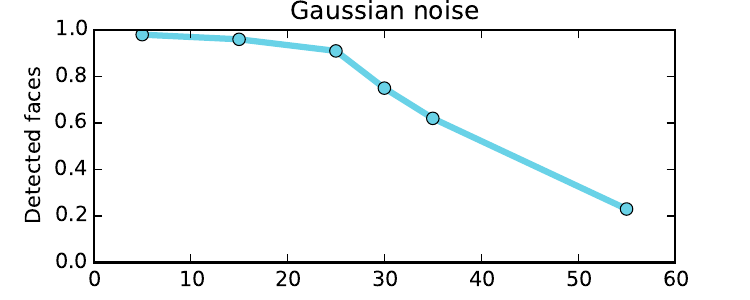}
\includegraphics[width=1\linewidth]{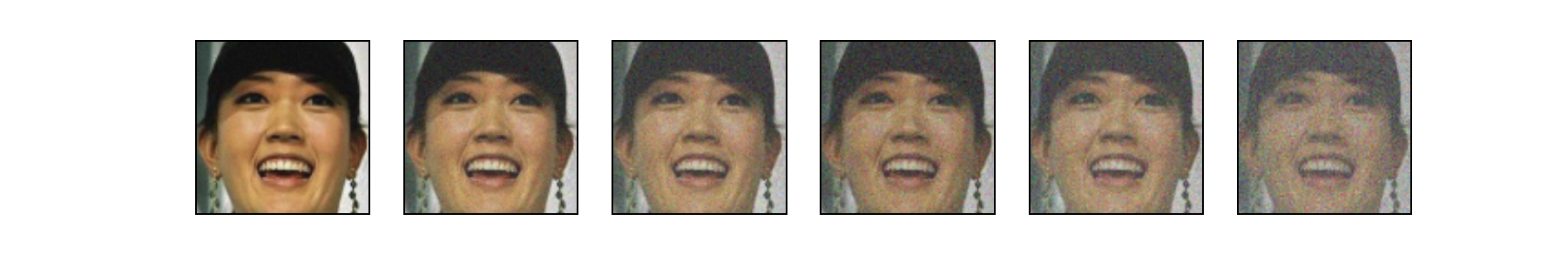}
\caption{\label{fig:gaussian}Face detection accuracy when applying random Gaussian noise to each pixel of the test image. Under the strongest noise settings ($\sigma$=50 intensity steps), Facebook's face detector finds 20\% of the test faces.}
\end{figure}

One simple filter perturbs each pixel's intensity by a distribution drawn from $N(0, \sigma)$. After perturbation, the dynamic range of the image is normalized to lie within 0-255. This tends to reduce the image's perceived contrast at high $\sigma$. Example images and results are shown in Fig.~\ref{fig:gaussian}.

This is one of our most promising \emph{machine-hurting} filters: under the highest noise setting, facebook can find less than 20\% of the faces in the image, but a human has little trouble detecting faces in the photograph. Our human visual system is robust to this kind of high-frequency noise, but Facebook's detector is not. This presents one easy way to hide faces that does not affect images very much.

\subsection{Lines through eyes}
\begin{figure}[t]
\includegraphics[width=1\linewidth]{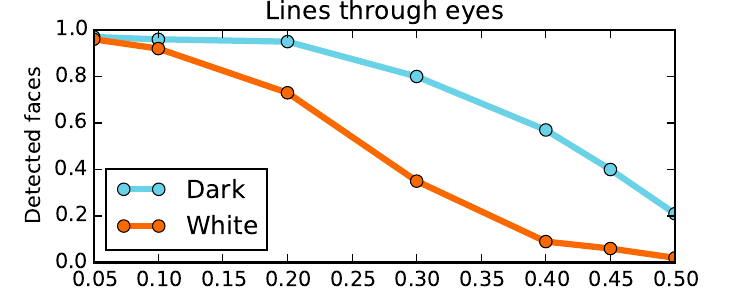}
\includegraphics[width=1\linewidth]{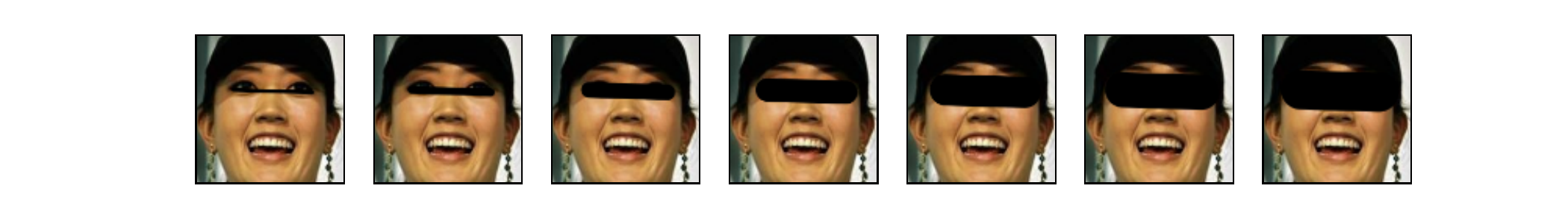}
\includegraphics[width=1\linewidth]{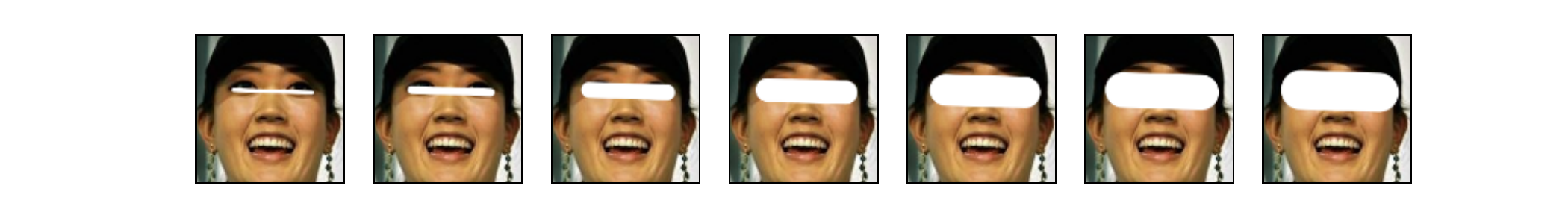}
\caption{\label{fig:lineeyes}Face detection accuracy when drawing white and black lines through the eyes, varying line thickness. This is the most effective filter methodology. Completely covering the periocular region in white successfully hides almost all faces from Facebook's face detector. Using a black line is not as effective.}
\end{figure}

Another \emph{machine-hurting} filter is to occlude the eyes by drawing ``celebrity censor bars'' over the periocular region. Pubfig does not include eye point locations, so for this technique, we used the publicly available implementation of Robust Cascaded Pose Regression~(RCPR)~\cite{burgos-artizzu_robust_2013} to estimate the locations of the outer and inner edges of the eyes and manually inspected the results for accuracy. We then drew a rounded line through these points and varied its thickness. Example images and performance results are shown in Fig.~\ref{fig:lineeyes}.

This classic censorship method is still reasonably effective. If the photographer obscures the periocular region with a white line, most faces in the dataset are not detected. One interesting finding is that \emph{color matters}: white occlusions are more effective than dark occlusions at hiding the face. This may be because face detectors can no longer exploit the discriminative dark shadows in the periocular region as a detection cue if it is colored white. 

\subsection{Darkening the image}
\begin{figure}[t]
\centering
\includegraphics[width=1\linewidth]{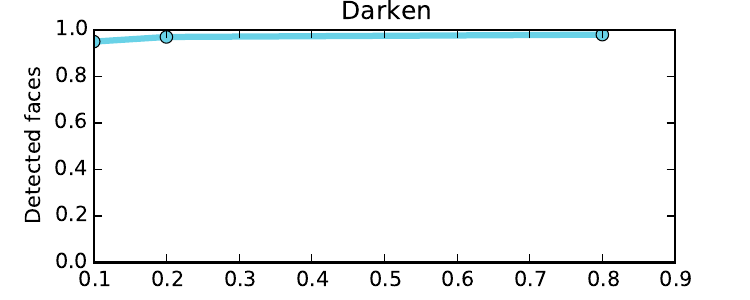}
\includegraphics[width=0.8\linewidth]{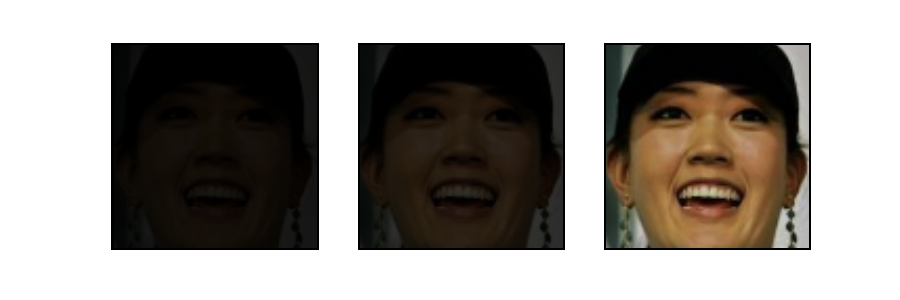}
\caption{\label{fig:darken}Face detection accuracy when darkening the entire image. Facebook performance does not degrade significantly, even if the image is almost entirely black.}
\end{figure}

Within the face detection pipeline, it is common to normalize the contrast of the image to correct pictures that are too dark or too bright. However, people have a hard time finding faces in uniformly dark images. This means darkening the image is not an effective way to hide faces -- it \emph{hurts human performance} without affecting Facebook's performance at all. Example images and face detection results are shown in Fig.~\ref{fig:darken}.

\subsection{Leopard spots}
\begin{figure}[t]
  \begin{minipage}{0.7\linewidth}
\includegraphics[width=1\linewidth,clip,trim=0mm 0mm 8mm 0mm]{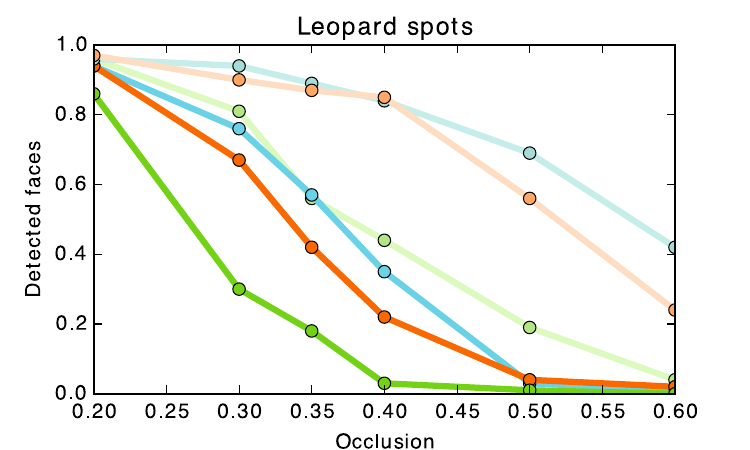}
\end{minipage}
\begin{minipage}{0.29\linewidth}
\newcommand{\circcolor}[1]{\tikz[baseline=-0.7ex]{\path[fill,color=#1] (0,0) circle (1ex);}}
\definecolor{69D2E7}{HTML}{69D2E7}
\definecolor{FA6900}{HTML}{FA6900}
\definecolor{73d216}{HTML}{73d216}
\definecolor{C7EBE8}{HTML}{C5EEEB}
\definecolor{ffB886}{HTML}{FFDDC2}
\definecolor{smallblurrycolor}{HTML}{DCF9BE}
\resizebox{1\linewidth}{!}{%
\begin{tabular}{|r@{}c@{\hskip 0.5ex}c|}
  \hline
&\textbf{Clear}&\textbf{Blurry}\\
\textbf{Big}&\circcolor{69D2E7}&\circcolor{C7EBE8}\\
\textbf{Med}&\circcolor{FA6900}&\circcolor{ffB886}\\
\textbf{Small}&\circcolor{73d216}&\circcolor{smallblurrycolor}\\
  \hline
\end{tabular}
}
\end{minipage}

\includegraphics[clip,trim=20mm 0mm 10mm 0mm, width=1.05\linewidth]{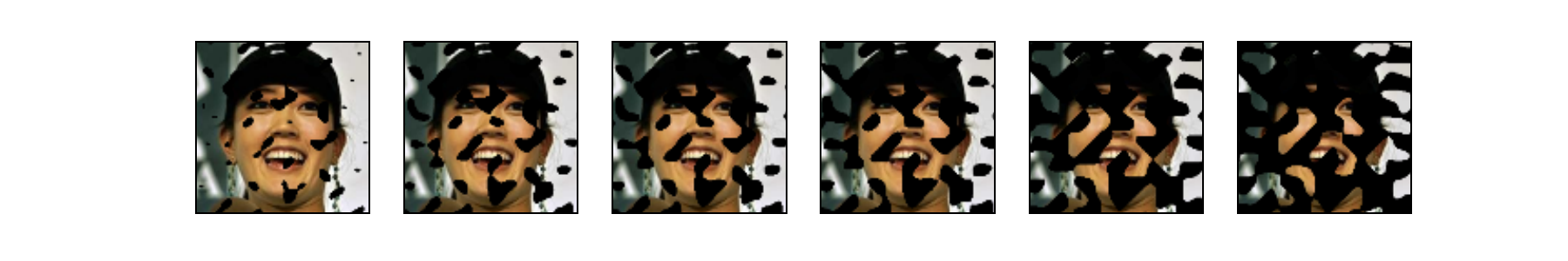}
\includegraphics[clip,trim=0mm 0mm 8mm 0mm, width=0.55\linewidth]{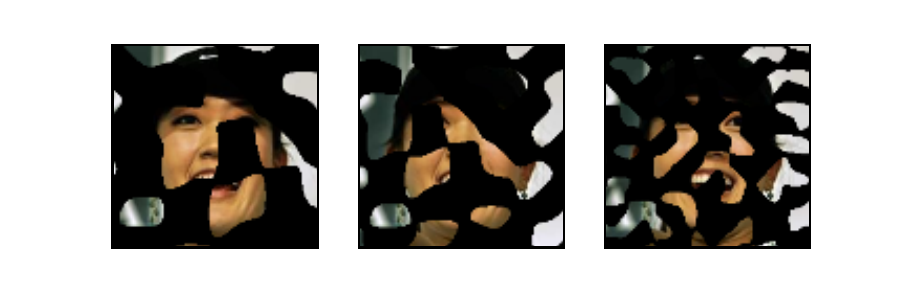}
\includegraphics[clip,trim=0mm 0mm 8mm 0mm, width=0.35\linewidth]{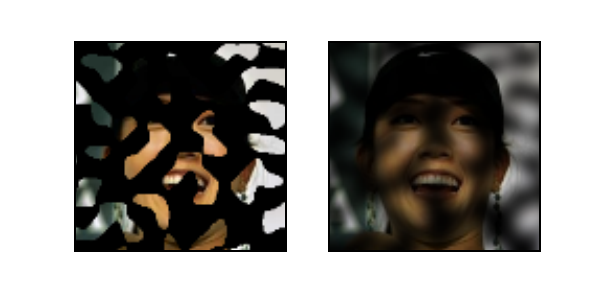}

\caption{\label{fig:leopard}Face detection accuracy when adding ``leopard spots'' (thresholded simplex noise). This plot varies the occluded area of the face (plot and top row of examples). We also varied the scale of the noise (\emph{Big}, \emph{Med}, and \emph{Small}; see bottom left examples), and we also applied Gaussian blur to the noise after thresholding (\emph{Clear} and \emph{Blurry}; see bottom right examples). The top row of examples are points along the dark green line and represent the hardest faces for Facebook to detect. }
\end{figure}

One last interesting method to hide faces is by adding ``leopard spots'' to the face. This is similar in spirit to Scheirer \etal's `` Face in the Branches'' test~\cite{scheirer_perceptual_2014}. First, we generate a mask by thresholding low-frequency 2D simplex noise. We can vary the amount of visible face area by changing the threshold. Since the size of Pubfig images varies widely, we rescale the frequency of the noise by the inter-ocular distance. We then set all masked face pixels to black. This adds low-frequency noise to the image, making it harder for machines and humans alike to find the face. We can further modify this filter by varying the scale of the noise and blurring the mask to add ``softness'' to the leopard spots. Example images and results are shown in Figure~\ref{fig:leopard}.

Adding leopard spots is another interesting \emph{machine-hurting} filter. Without blur, covering 40\% of the face area with small leopard spots makes almost all of the faces completely invisible to Facebook's face detector, but humans can still see the face area reasonably well. However, some parameters are not effective. For instance, increasing the scale of the noise and adding more blur still leaves most faces detectable.

\section{Natural occlusions}
The above synthetic experiments help us understand what kinds of image transformations that Facebook's face detector is invariant to, but they require the uploader to modify the image. If the image was captured unintentionally or in a surveillance scenario, it is not possible for the privacy-conscientious subject to cooperate with the image uploader. In this section, we investigate steps the \emph{subject themselves} can take to obscure their face.

\subsection{Occluding clothing}\label{sec:occluding-clothing}
\begin{figure}[t]
  \includegraphics[width=1\linewidth]{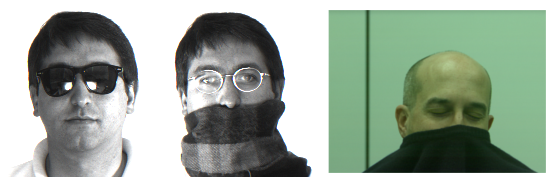}
  \caption{\label{fig:arface}Examples of images appearing in ARFace (left, middle) and UMB (right). ``Scarf'' photos in UMB are more likely to contain scarves that cover the subject's nose, and the subject's eyes are closed more often.}
\end{figure}

One simple approach that will not raise suspicion is to wear occluding clothing such as a scarf, hat, or big sunglasses. To measure the effectiveness of these simple evasive measures, we evaluate Facebook's detection performance on two databases. Example images are shown in Fig.~\ref{fig:arface}.

The classic ARFace dataset~\cite{arface}, now 18 years old, contains carefully captured pictures of 135 subjects obtained under a controlled laboratory setting. For our setting, we use 417 ``neutral'' images (subsets 1, 5, and 6), 417 images of subjects wearing scarves (subsets 11, 12, 13), and 417 images of subjects with sun glasses (subsets 8, 9, and 10).

We also evaluate performance on the UMB-DB 3D face dataset~\cite{umb-db}, which contains 2D and 3D captures of 143 subjects. Each image is tagged with several binary attributes such as ``scarf,'' ``smile,'' ``free,'' and ``occluded.''

\begin{table}[t]
\begin{tabular}{l r r}
  Dataset&Number&Detection probability\\
\hline
\textbf{ARFace}\cite{arface}\\
Neutral           &    417 &    0.995 \\
Wearing scarves   &    417 &    0.880 \\
Wearing glasses   &    417 &    0.964 \\
\hline
\textbf{UMB-DB}\cite{umb-db}\\
Neutral           &    883 &    0.998 \\
Wearing scarf     &    151 &    0.570 \\
Occluding haircut &     33 &    0.909 \\
Hat               &    183 &    0.945 \\
\end{tabular}
\caption{\label{tab:scarf-results}Facebook's detection probability on two datasets of face images with naturally-occluding clothing.}
\end{table}

Detection results are shown in Tab.~\ref{tab:scarf-results}. Generally, no occluding clothing is completely effective in hiding the face from Facebook's face detector. Wearing a scarf appears to be the best way to avoid detection, lowering the detection probability to 88\% (ARFace) or 57\% (UMB-DB). We hypothesize two reasons for this potential difference: in ARFace, the scarves worn by the subjects only cover the mouth region of the face and do not typically cover the nose. In the UMB set, images with the `scarf' label feature scarves that typically cover the subject's nose. Second, subjects in UMB `scarf' images typically close their eyes while ARFace subjects usually keep their eyes open.


\subsection{Keypoint occlusion}
To gain a more detailed look at which regions of the face cause the most impact on detection rates, we evaluated Facebook's detection performance on the ``Caltech Occluded Faces in the Wild'' (COFW)~\cite{rcpr} training set, which contains 1345 unconstrained images and keypoint annotations. Each keypoint has a binary ``is-occluded?''\ flag.

\begin{figure}[t]
\includegraphics[width=1\linewidth]{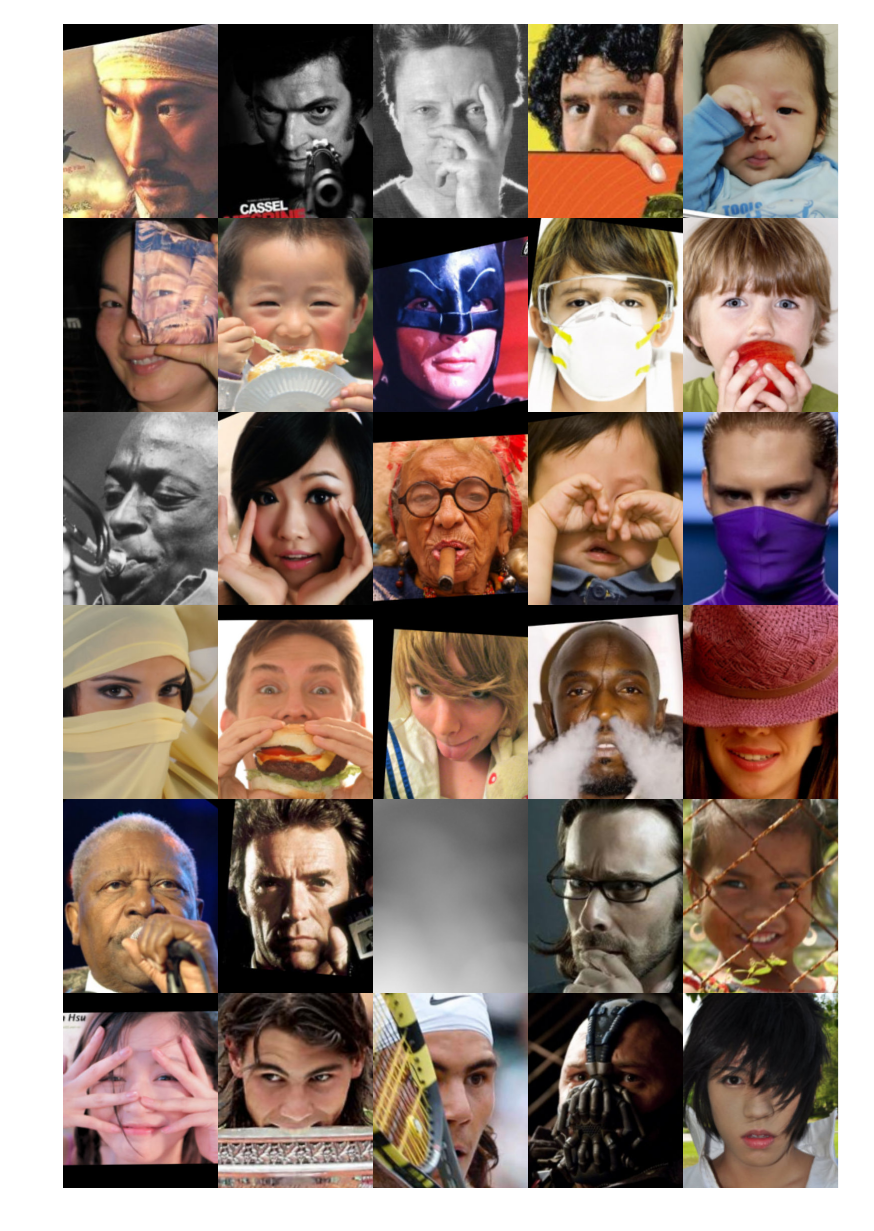}
\caption{\label{fig:cofw-failures}All 30 missed faces in the COFW training set.}
\end{figure}

Facebook's face detector detects 97.8\% of the faces in the COFW training set. All 30 missed faces are shown in Fig.~\ref{fig:cofw-failures}. Since this dataset was constructed to feature mostly occlusions, it is not surprising that all but two of the missed faces contain heavy occlusions. However, we can get a sense for what kinds of occlusions are most likely to cause missed detections by computing detection statistics for each keypoint.

\begin{figure}[t]
  \centering
  \includegraphics[width=0.7\linewidth]{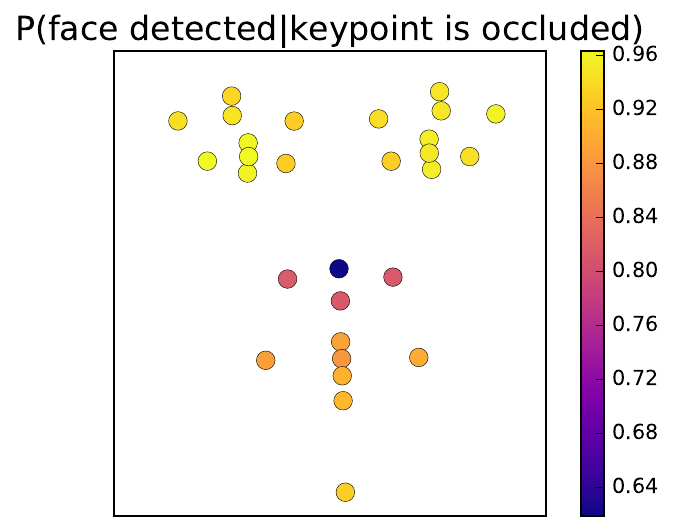}
  \caption{\label{fig:cofw-keypoint-distributions} Figure showing the probability that a face is detected given a certain keypoint is occluded. Example: only 61.9\% of the faces with covered nose tips are detected.}
\end{figure}

The color of each keypoint in Fig.~\ref{fig:cofw-failures} shows face detection probability among images where that keypoint is occluded. From this figure, we can see that Facebook's face detector will only find 60\% of faces that have the nose tip occluded. This is likely because objects that cover the nose tip often occlude many more points as well. Further, covering the mouth region lowers detection probability more than covering the eye regions. These results, and those in \ref{sec:occluding-clothing}, seem to indicate that covering the nose and mouth regions may be a reasonable way to hide from some face detectors, but more exploration is necessary.

\section{Future work}
Thus far, this work focused exclusively on the \emph{detection} step. However, we know that it is possible to use the social network's recognition system in interesting ways. In future work, we want to uncover various privacy abuses, such as:
\begin{itemize}
\item \textbf{Using face recognition in social networks to build an automatic ``dragnet.''} Since conventional social networks such as Facebook can typically recognize anyone on a user's buddy list, this explicit social link in fact grants \emph{implicit} access to that user's recognition model. This means that under default privacy settings, users can use the tools presented here to automatically recognize their friends in any photos of their choice. This can be abused as follows: Suppose Louise sets up a surveillance camera in front of her house. Whenever the camera detects motion, a script uploads a few frames of camera footage to Facebook and downloads classification results. The script sends Louise a text message when any of her friends arrive. In this way, Louise gains the ability to know whenever \emph{anyone} on her buddy list visited her house. This is interesting because
Louise does not need to collect any kind of training set. Facebook \emph{automatically and implicitly} provides her with the ability to query near-perfect, up-to-date models of many of her friends, giving her a state-of-the-art recognition framework without the hassle of enrollment or parameter validation. Her friends' models improve over time as they continue to use the website and upload more pictures.

\item \textbf{Misleading recognition results.} Is it possible to poison one's model to classify someone else? If Sara uploads several pictures of Mason to her profile and tags herself as Mason's face, social network face recognition systems might become confused and begin to automatically tag pictures of Mason as Sara. This would notify Sara whenever any of her friends are in photos of Mason.

\end{itemize}

\section{Conclusion}
Though we only evaluated Facebook's face detector, our results could apply to any social medium that uses automatic face recognition technology. Our goal is not to complain about Facebook's default policy settings or lament the impending death of privacy. Instead, we wish to start a discussion about the unintended power dynamics that these systems create when they cannot be avoided. After all, avoiding automatic face detection and recognition is becoming more difficult as talented engineers search for ways to improve these systems. Real-world countermeasures such as the privacy visor are not always effective and may even make the subject \emph{more} recognizable in the real world.

In this work, we studied image transformation techniques that help a privacy-conscious individual avoid being automatically identified. However, there are several practical problems with the methods we outline. First, the \emph{photo uploader}, not the individual, must remember to use the image perturbation techniques. Second, many of these techniques \emph{make the image look worse} to humans. However, these techniques illuminate the strengths and weaknesses of state-of-the-art face detectors used in common social network platforms. We now know that Facebook has little trouble detecting faces in low-light conditions, but occlusions and noise are still difficult to find. If a privacy-seeking individual wishes to develop more ways of avoiding automatic detection, building from these observations could be a good first start.

\bibliographystyle{ieee}
\bibliography{sp-sources}

\begin{thebibliography}{10}\itemsep=-1pt

\bibitem{babenko-miltrack}
B.~Babenko, M.-H. Yang, and S.~Belongie.
\newblock Robust object tracking with online multiple instance learning.
\newblock {\em Pattern Analysis and Machine Intelligence, {IEEE} Transactions
  on}, 33(8):1619--1632, 2011.

\bibitem{belhumeur_localization}
P.~Belhumeur, D.~Jacobs, D.~Kriegman, and N.~Kumar.
\newblock Localizing parts of faces using a consensus of exemplars.
\newblock In {\em {CVPR}}, pages 545--552, June 2011.

\bibitem{berg_tom-vs-pete_2012}
T.~Berg and P.~N. Belhumeur.
\newblock Tom-vs-pete classifiers and identity-preserving alignment for face
  verification.
\newblock In {\em {BMVC}}, 2012.

\bibitem{bruce_whats_1994}
V.~Bruce, M.~A. Burton, and N.~Dench.
\newblock What's distinctive about a distinctive face?
\newblock {\em The Quarterly Journal of Experimental Psychology Section A},
  47(1):119--141, Feb. 1994.

\bibitem{rcpr}
X.~Burgos-Artizzu, P.~Perona, and P.~Dollar.
\newblock Robust {Face} {Landmark} {Estimation} under {Occlusion}.
\newblock In {\em {ICCV} {W}orkshops}, pages 1513--1520, Dec. 2013.

\bibitem{burgos-artizzu_robust_2013}
X.~P. Burgos-Artizzu, P.~Perona, and P.~Dollár.
\newblock Robust face landmark estimation under occlusion.
\newblock In {\em {ICCV}}, 2013.

\bibitem{umb-db}
A.~Colombo, C.~Cusano, and R.~Schettini.
\newblock {UMB}-{DB}: {A} database of partially occluded 3d faces.
\newblock In {\em {ICCV Workshops}}, pages 2113--2119, Nov. 2011.

\bibitem{goldman_facebook_2015}
D.~Goldman.
\newblock Facebook is scanning your photos to help you share them.
\newblock {\em CNN Money}, Nov. 2015.

\bibitem{lfw-results}
G.~B. Huang, M.~Ramesh, T.~Berg, and E.~Learned-Miller.
\newblock {LFW} results.
\newblock Internet: \url{http://vis-www.cs.umass.edu/lfw/results.html}.
\newblock [Accessed: Sep. 25, 2015].

\bibitem{juefei-xu_preliminary_2015}
F.~Juefei-Xu, D.~Pal, K.~Singh, and M.~Savvides.
\newblock A {Preliminary} {Investigation} on the {Sensitivity} of {COTS} {Face}
  {Recognition} {Systems} to {Forensic} {Analyst}-{Style} {Face} {Processing}
  for {Occlusions}.
\newblock In {\em {CVPR}}, pages 25--33, 2015.

\bibitem{attribute_and_simile}
N.~Kumar, A.~C. Berg, P.~N. Belhumeur, and S.~K. Nayar.
\newblock {A}ttribute and {S}imile {C}lassifiers for {F}ace {V}erification.
\newblock In {\em {ICCV}}, Oct 2009.

\bibitem{arface}
A.~M. Martinez and R.~Benavente.
\newblock The {AR} {F}ace database.
\newblock {\em CVC Technical Report}, 24, 1998.

\bibitem{fedhd}
J.~Parris, M.~Wilber, B.~Heflin, H.~Rara, A.~El-Barkouky, A.~Farag,
  J.~Movellan, M.~Castrilon-Santana, J.~Lorenzo-Navarro, M.~Teli, S.~Marcel,
  C.~Atanasoaei, and T.~Boult.
\newblock Face and eye detection on hard datasets.
\newblock In {\em 2011 International Joint Conference on Biometrics ({IJCB})},
  2011.

\bibitem{ren_face_2014}
S.~Ren, X.~Cao, Y.~Wei, and J.~Sun.
\newblock Face alignment at 3000 {FPS} via regressing local binary features.
\newblock In {\em {CVPR}}, pages 1685--1692. IEEE, 2014.

\bibitem{sarno_attractiveness_1997}
J.~A. Sarno and T.~R. Alley.
\newblock Attractiveness and the memorability of faces: Only a matter of
  distinctiveness?
\newblock {\em The American Journal of Psychology}, 110(1):81--92, Apr. 1997.

\bibitem{tboult}
W.~Scheirer and T.~Boult.
\newblock Biometrics: Practical issues in privacy and security.
\newblock Technical report, Securics, Inc. and the University of Colorado
  Colorado Springs, 2011.
\newblock Tutorial at International Jount Conference on Biometrics.

\bibitem{scheirer_perceptual_2014}
W.~J. Scheirer, S.~E. Anthony, K.~Nakayama, and D.~D. Cox.
\newblock Perceptual annotation: Measuring human vision to improve computer
  vision.
\newblock {\em IEEE Transactions on Pattern Analysis and Machine Intelligence
  (T-PAMI)}, 36, August 2014.

\bibitem{sun_deepid2}
Y.~Sun, X.~Wang, and X.~Tang.
\newblock Deep learning face representation by joint
  identification-verification.
\newblock {\em {arXiv}:1406.4773 [cs]}, June 2014.
\newblock {arXiv}: 1406.4773.

\bibitem{taigman_deepface:_2013}
Y.~Taigman, M.~Yang, M.~Ranzato, and L.~Wolf.
\newblock Deepface: Closing the gap to human-level performance in face
  verification.
\newblock In {\em {CVPR}}, 2013.

\bibitem{viola-facedetection}
P.~Viola and M.~Jones.
\newblock Robust real-time face detection.
\newblock {\em International Journal of Computer Vision ({IJCV})},
  57(2):137--154, May 2004.

\bibitem{wang_improving_2008}
H.~Wang and J.~Chen.
\newblock Improving self-quotient image method of {NPR}.
\newblock In {\em 2008 International Conference on Computer Science and
  Software Engineering}, volume~6, pages 213--216, 2008.

\bibitem{wang_face_2004}
H.~Wang, S.~Li, and Y.~Wang.
\newblock Face recognition under varying lighting conditions using self
  quotient image.
\newblock In {\em Sixth {IEEE} International Conference on Automatic Face and
  Gesture Recognition, 2004. Proceedings}, pages 819--824, 2004.

\bibitem{yamada_privacy_2013}
T.~Yamada, S.~Gohshi, and I.~Echizen.
\newblock Privacy {Visor}: {Method} for {Preventing} {Face} {Image} {Detection}
  by {Using} {Differences} in {Human} and {Device} {Sensitivity}.
\newblock In B.~D. Decker, J.~Dittmann, C.~Kraetzer, and C.~Vielhauer, editors,
  {\em Communications and {Multimedia} {Security}}, number 8099 in Lecture
  {Notes} in {Computer} {Science}, pages 152--161. Springer Berlin Heidelberg,
  Sept. 2013.
\newblock DOI: 10.1007/978-3-642-40779-6\_13.

\end{thebibliography}







\end{document}